\documentclass{article}
\usepackage{ijcai26}

\usepackage{times}
\usepackage{soul}
\usepackage{url}
\usepackage[hidelinks]{hyperref}
\usepackage[utf8]{inputenc}
\usepackage[small]{caption}
\usepackage{graphicx}
\usepackage{amsmath}
\usepackage{amsthm}
\usepackage{booktabs}
\usepackage{algorithm}
\usepackage{algorithmic}
\usepackage[switch]{lineno}
\usepackage{amssymb}
\usepackage{multirow}
\usepackage{siunitx}

\title{FusionCast: Enhancing Precipitation Nowcasting with Asymmetric Cross-Modal Fusion and Future Radar Priors}

\author{
Henan Wang$^1$
\and
Shengwu Xiong$^1$\and
Yifang Zhang$^1$\and
Wenjie Yin$^2$\and
Chen Zhou$^2$\and
Yuqiang Zhang$^2$\And
Pengfei Duan$^1$\thanks{Corresponding author.}\\
\affiliations
$^1$School of Computer Science and Artificial Intelligence, Wuhan University of Technology\\
$^2$School of Earth and Space Science and Technology, Wuhan University\\
}
\begin{document}

\maketitle

\begin{abstract}
    Deep learning has significantly improved the accuracy of precipitation nowcasting. However, most existing multimodal models typically use simple channel concatenation or interpolation methods for data fusion, which often overlook the feature differences between different modalities. This paper therefore proposes a novel precipitation nowcasting optimisation framework called FusionCast. This framework incorporates three types of data: historical precipitable water vapour (PWV) data derived from global navigation satellite system (GNSS) inversions, historical radar-based quantitative precipitation estimation (QPE), and forecasted radar QPE serving as a future prior. The FusionCast model comprises two core modules: the future prior radar QPE processing Module, which forecasts future radar data; and the Radar–PWV Fusion (RPF) module, which uses a gate mechanism to efficiently combine features from various sources. Experimental results show that FusionCast significantly improves nowcasting performance.
\end{abstract}

\section{Introduction}
    Against the background of global warming, the frequency and intensity of extreme weather events have increased significantly, presenting unprecedented challenges to human society. According to the latest statistics from the World Meteorological Organization (WMO), over the 50-year period from 1970 to 2021, a total of 11,778 disasters caused by extreme weather, climate, and water-related events were reported worldwide.  These disasters resulted in over \textbf{2 million} fatalities and resulted in staggering economic losses amounting to US\$4.3 trillion~\cite{wmo:intro}. In response to such disasters, precipitation nowcasting is defined as the detailed forecasting of weather conditions with high spatio-temporal resolution for the next 0-6 hours~\cite{whatextreme}, which is crucial for enhancing the time efficiency and accuracy of disaster warnings.
    
    Doppler weather radar has long been a key data source for nowcasting because of its high spatio-temporal resolution~\cite{doppler}. Early methods mainly use opticalflow-based echo extrapolation and assumed steady motion over short periods~\cite{pysteps}, but they often ignore rainfall growth and decay, which limits their ability to handle complex convective processes~\cite{nwplimiation}.
    
    Recent progress in deep learning has driven rapid development in precipitation nowcasting, from early convolutional long short-term memory (ConvLSTM) and U-Net models~\cite{convolutional,ayzel2020rainnet} to methods based on Generative Adversarial Networks (GANs) that offer probabilistic forecasts and remain effective at longer lead times and higher resolutions~\cite{nowcastnet}. More recently, diffusion models have been introduced to generate realistic rainfall fields~\cite{cascast}, further showing that data-driven methods can capture spatiotemporal patterns and improve forecast accuracy.

    As more observations become available, fusing radar, satellite, and GNSS data has become a common way to improve precipitation forecasts, since radar alone cannot fully describe rainfall~\cite{radarlimiation1}. However, most multimodal models still rely on simple channel stacking or interpolation~\cite{fusion1}, even though radar QPE and PWV are fundamentally different, one being a dense field and the other a point-based vapour measurement.~\cite{gnss}. Ignoring these differences often leads to feature interference and weaker predictions.

    In order to address the limitations inherent in fusion mechanisms, a novel fusion framework is proposed. Called FusionCast, this framework is capable of perceiving modal characteristics. Unlike traditional approaches, FusionCast integrates three key modalities through differentiated processing mechanisms: historical PWV (representing water vapour conditions), historical radar QPE (representing evolution trends), and generated future a priori radar QPE (representing positional a priori information).

    Two core modules have been developed in order to enhance forecasting accuracy. Firstly, a future prior generation module utilises high-performing single-modal models to generate future radar echo fields as positional guidance. Secondly, in order to overcome feature misalignment, which is inherent in conventional concatenation, RPF module has been innovatively introduced. This module does not simply superimpose characteristics, instead dynamically perceiving the efficacy of different modalities through gating units. The model balances GNSS-derived water vapor information with radar observations according to the current precipitation pattern, and uses future radar guidance only when it is reliable. In this way, the model does not simply take all inputs at face value, but instead focuses on the information that is most helpful for the current forecast.

    The main contributions of this paper are summarised as follows:
    \begin{itemize}
        \item We propose FusionCast. It improves precipitation nowcasting accuracy through differentiated modelling of heterogeneous modalities.
        \item We design the RPF module, which effectively addresses the challenges of feature alignment and fusion across multimodal inputs.
        \item We empirically demonstrate that incorporating a prior modality significantly enhances precipitation nowcasting performance, and validate the effectiveness of the proposed prior-based strategy on the Mississippi River Basin (MRB) dataset.
    \end{itemize}
    \nocite{dgmr}
    \nocite{diffcast}
    \nocite{prediff}
    \nocite{probabilistic:cit}
    \nocite{lu2025rsg}
    \nocite{pwv}

\section{Related Work}
    Traditional precipitation nowcasting primarily relies on optical flow techniques and numerical weather prediction (NWP). However, traditional precipitation nowcasting methods lack sufficient accuracy in capturing short-duration intense convective processes and extreme weather events, and involve high computational costs~\cite{NWP}.

    In recent years, deep learning has become the dominant approach in this area, largely because of its strong ability to model complex nonlinear relationships. The ConvLSTM model was one of the first to provide a widely used baseline for spatio-temporal sequence prediction~\cite{convlstm}. Building on this, TrajGRU~\cite{trajgru} was proposed to better handle motion variation that standard convolutional recurrent networks struggle to represent. To reduce the tendency of predicted images to become overly smooth, PredRNN~\cite{wang2022predrnn} and its variants further improved spatial memory by introducing spatio-temporal LSTM structures.

    More recently, cGANs and diffusion models have been used for precipitation nowcasting and have improved the visual quality of high-resolution forecasts~\cite{cGAN}. NowcastNet blends physical motion constraints with adversarial training to better follow radar echo evolution while keeping the results sharp, and CasCast uses a cascaded diffusion design to better capture fine-scale structures~\cite{stablediffusion}.
    % More recently, conditional generative adversarial networks (cGANs) and diffusion-based models have been applied to precipitation nowcasting, leading to noticeable improvements in the visual quality of high-resolution forecasts~\cite{cGAN}. NowcastNet combines physical evolution constraints with adversarial learning, which allows it to track complex radar echo motion while keeping the predictions sharp. CasCast introduces a cascaded diffusion framework, in which a staged modelling strategy is used to capture fine-scale atmospheric structures more effectively~\cite{stablediffusion}.

    Although these methods have improved radar echo extrapolation, most still rely only on past radar data, which limits their ability to capture complex atmospheric behaviour. Multimodal fusion alleviates this by bringing together different precipitation-related sources, and recent deep learning models have made this integration more effective~\cite{hu2025multisource}.
    % Although these methods have led to clear improvements in radar echo extrapolation, most of them still depend only on past radar observations, which makes it hard to fully represent complex atmospheric behaviour. Multi-modal data fusion addresses this limitation by combining different sources of precipitation-related information to improve forecasting performance. With the development of deep learning, such fusion networks have been increasingly used to better reflect the physical complexity of the atmosphere~\cite{sca}.

    MetNet is a representative line of work from Google Research. The original version used axial self-attention to combine radar and satellite images and capture large-scale spatial patterns~\cite{metnet1}. MetNet-2 expanded the receptive field to support forecasts up to 12 hours~\cite{metnet2}. MetNet-3 further brought in station observations alongside gridded inputs, showing that point measurements help correct and constrain grid-based forecasts~\cite{metnet3}.
    % The MetNet series from Google Research is a representative example in this line of work. The original MetNet introduced an axial self-attention mechanism to jointly process radar data and satellite cloud images, allowing the model to capture large-scale spatial context. MetNet-2 extended this design by enlarging the receptive field, enabling precipitation forecasts up to 12 hours ahead. More recently, MetNet-3 explicitly focuses on data heterogeneity by encoding sparse station measurements together with dense gridded inputs, highlighting the importance of physical point observations for constraining and correcting grid-based predictions.

    However, existing multimodal fusion architectures face two primary challenges:
    \begin{itemize}
        \item \textbf{Simple Fusion Strategies:} Most methods achieve multimodal fusion by concatenating channels or interpolating sparse site data spatially. However, such strategies often overlook significant differences between data sources, which can undermine the complementary advantages of multimodal information.
        \item \textbf{Lack of prior information:} The future outputs of radar forecasts themselves progressively attain high reliability and are capable of reflecting the evolution of precipitation systems. However, few studies have systematically explored how to use the future outputs of existing high-performance prediction models as prior information to further enhance forecast results.
    \end{itemize}
    Departing from the fusion methods mentioned above, FusionCast employs the RPF module to adaptively integrate multimodal features, while the prior module leverages future forecasts to improve nowcasting accuracy.
\nocite{WRF}
\nocite{fsrgan}
\nocite{bhuskute2025multisource}
\nocite{chen2025fuxi}

\section{Methodology}
    \begin{figure*}[htbp]
        \centering
        \includegraphics[width=0.85\textwidth]{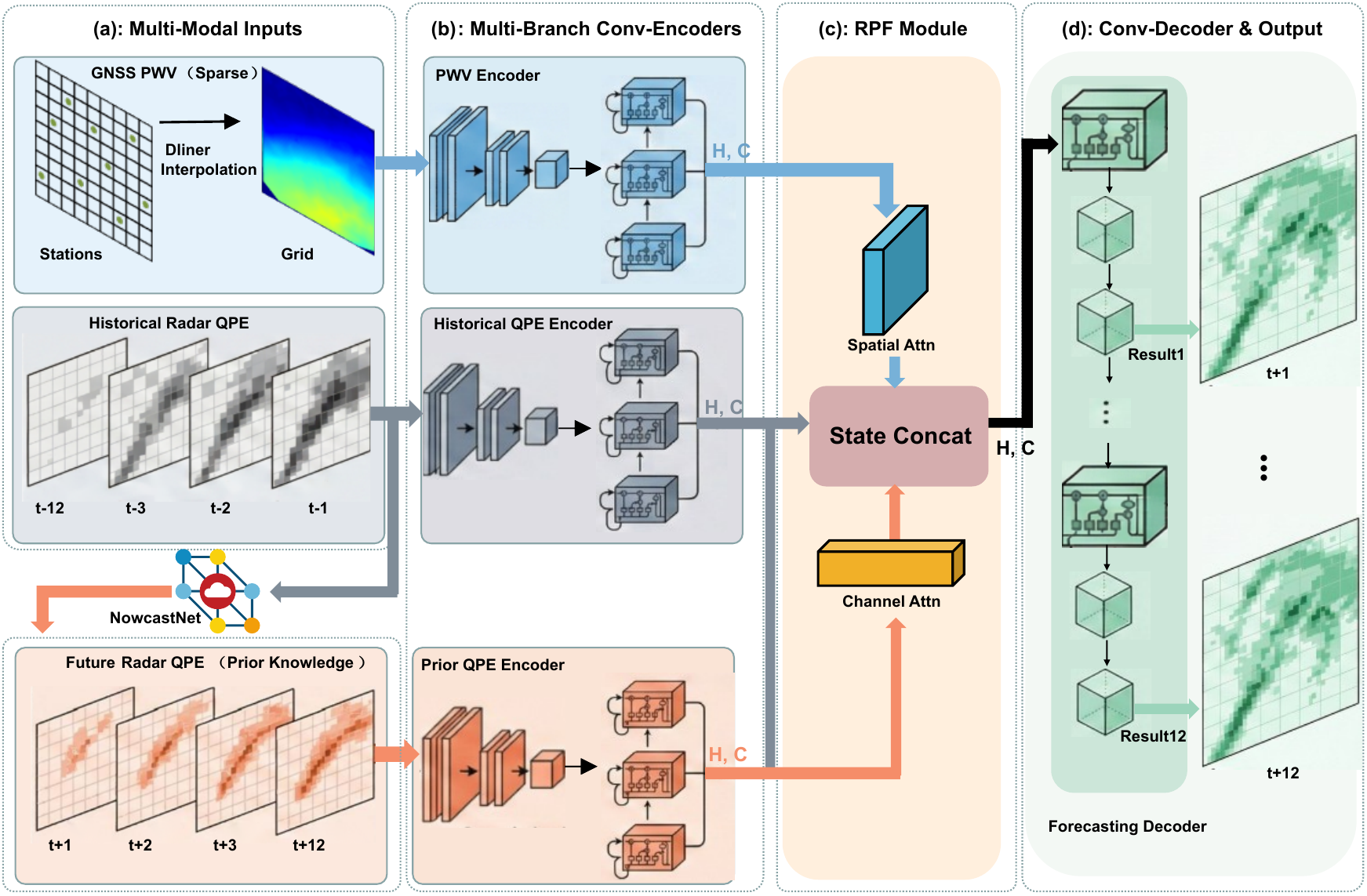}
        \caption{(a) Incorporates sparse GNSS PWV stations, historical radar QPE data, and future radar QPE data as model inputs. (b) Different data types are encoded separately through a multi-branch architecture. (c) Information from diverse data sources is fused to enhance forecasting performance. (d) The final decoder generates forecast outputs, providing projections of future precipitation.}
        \label{fig: FusionCast}
    \end{figure*}
    In this study, we propose a novel deep learning framework, termed \textbf{FusionCast}, for high-resolution precipitation nowcasting. The architecture is designed to effectively fuse multimodal data sources: sparse PWV fields, high-resolution historical radar QPE, and future prior knowledge derived from the NowcastNet model. The framework adopts an Encoder-Decoder structure integrated with a spatial channel attention mechanism to maximize the complementary advantages of different modalities.The overall framework is illustrated in Figure~\ref{fig: FusionCast}.
    \subsection{Overview of the Architecture}
        The model consists of three parallel encoding branches and one decoding branch based on the ConvLSTM unit, which is capable of capturing spatiotemporal correlations. The inputs to the model are defined as follows:
        \begin{itemize}
            \item \textbf{PWV Branch}: Takes the sequence of interpolated PWV grids, denoted as $X_{PWV}$.
            \item \textbf{Historical Radar Branch}: Takes the observed historical radar QPE frames, denoted as $X_{Radar}^{hist}$.
            \item \textbf{Prior Knowledge Branch}: Takes the future radar extrapolation frames predicted by the NowcastNet model, denoted as $X_{Radar}^{prior}$.
        \end{itemize}
        The core innovation lies in the fusion strategy, where the hidden states extracted from these branches are merged via the RPF module to initialise the forecasting decoder.
    \subsection{Multimodal Encoders}
        To extract hierarchical spatiotemporal features, we employ three separate encoder networks.
        For the PWV and Historical Radar branches, the input sequences are first processed by two layers of Time-Distributed 2D Convolutions with a stride of 2. This downsampling operation reduces the spatial dimension to $1/4$ of the original input, expanding the receptive field and reducing computational cost. The downsampled feature maps are then fed into a ConvLSTM layer to encode temporal dynamics:
        \begin{equation}
            \begin{aligned}
                H_{PWV}, C_{PWV} &= \text{ConvLSTM}_{pwv}(\text{Conv2D}(X_{PWV})) \\
                H_{Hist}, C_{Hist} &= \text{ConvLSTM}_{hist}(\text{Conv2D}(X_{Radar}^{hist}))
            \end{aligned}
        \end{equation}
        where $H$ and $C$ represent the hidden state and cell state of the LSTM units, respectively.

        Simultaneously, the Prior Knowledge Branch processes the NowcastNet predictions. Given that these frames contain crucial information about future storm movement, we employ a deeper feature extraction with increased filter numbers (from 32 to 128) to capture richer semantic contexts:
        \begin{equation}
            \resizebox{.91\linewidth}{!}{$
            \displaystyle
                H_{Prior}, C_{Prior} = \text{ConvLSTM}_{prior}(\text{Conv2D}(X_{Radar}^{prior}))
            $}
        \end{equation}
    \subsection{RPF Module}
\begin{figure}[htbp]
            \centering
            \includegraphics[width=0.42\textwidth]{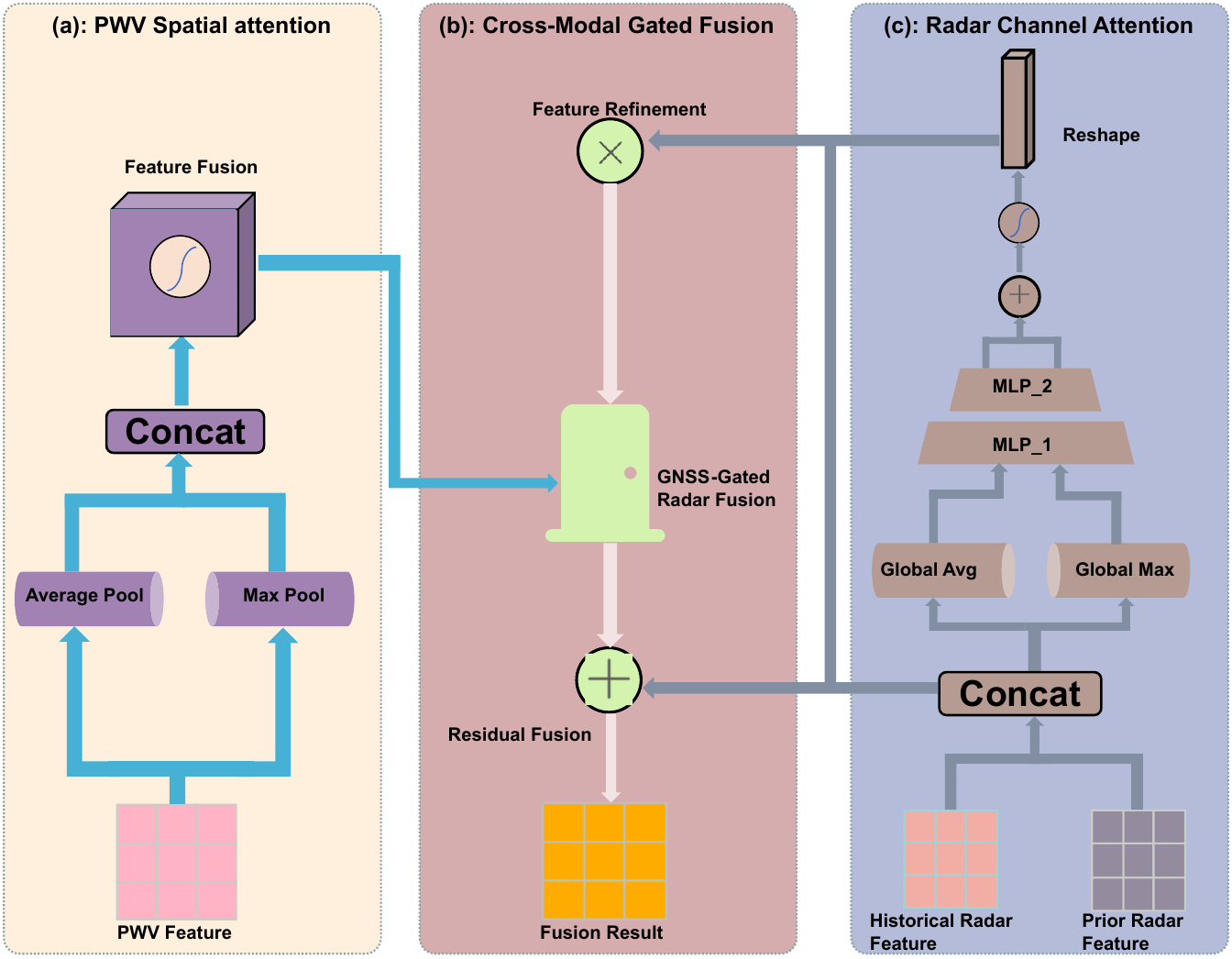}
            \caption{(a) A spatial attention mechanism is applied to PWV features to emphasise critical spatial information. (b) The integration of GNSS and radar data through feature fusion via a gating mechanism is demonstrated. (c) Channel attention mechanisms extract key features from radar data to optimise information representation.}
            \label{fig: RPF}
        \end{figure}

        We propose the RPF module to effectively fuse sparse PWV data into dense radar echo characteristics. The purpose of this model is to learn only incrementally validated information through PWV physical verification, thereby retaining the high-resolution texture of radar data while strictly adhering to atmospheric physics constraints. This module comprises two main sub-processes: PWV spatial attention and radar channel attention. Figure~\ref{fig: RPF} illustrates the RPF model architecture.
        \subsubsection{PWV spatial attention}
            We treat PWV characteristics ($F_{PWV}$) as the primary factor. To extract the most important features of the water vapour conditions, we first aggregate the features along the channel dimension to generate spatial descriptors using mean and max pooling. We then use the large receptive field of convolutional layers to capture continuous atmospheric water vapour fields. The spatial gate map ($M_{spatial}(F_{PWV})$) is ultimately generated via the sigmoid activation function:
            \begin{equation}
                \resizebox{.91\linewidth}{!}{$
                \displaystyle
                    M_{spatial}(x) = \sigma(\mathcal{f}^{7 \times 7}([\mathbf{AvgPool}_c(x); \mathbf{MaxPool}_c(x)]))
                $}
            \end{equation}
            where $\sigma$ denotes the sigmoid function, and $\mathcal{f}^{7 \times 7}$ represents the convolution operation. 
            From a physical perspective, $M_{spatial}$ represents not merely a spatial mask, but rather the \textbf{distribution of thermal energy} within the prevailing atmospheric conditions. It indicates which spatial regions possess the energy conditions necessary for precipitation formation, thereby providing a physically motivated admission criterion for subsequent radar features.
        \subsubsection{Radar channel attention}
            The radar echo $F_{radar}$ contains intricate textural information. To improve the discriminability of features, we aggregate spatial information using both global average and global max pooling operations ($\mathcal{P}_{avg}, \mathcal{P}_{max}$), followed by a shared MLP (Multi-Layer Perceptron), The channel weight function is defined as follows:
            \begin{equation}
                \resizebox{.91\linewidth}{!}{$
                \displaystyle
                    W_{channel}(x)=\sigma(MLP(\mathcal{P}_{avg}(x))+MLP(\mathcal{P}_{max}(x)))
                $}
            \end{equation}
            
            Subsequently, the Radar features are reweighted using this weighting factor to obtain the channel-refined features: $F'_{radar} = F_{radar} \otimes W_{channel}(F_{radar})$. This process enhances the \textbf{signal-to-noise ratio} (SNR) within the feature space, suppressing redundant channels that contribute little to precipitation forecasting while highlighting feature responses containing key echo textures.
        \subsubsection{Cross-Modal Gated Fusion}
            To avoid introducing noise caused by direct superposition, we apply the PWV-generated spatial gate $M_{spatial}(F_{PWV})$ to the refined radar feature $F'_{radar}$ to perform feature correction:
            \begin{equation}
                H_{Fused} = M_{spatial}(F_{PWV}) \otimes F'_{Radar} + F_{Radar}
            \end{equation}

            From a physical perspective, $M_{spatial}(F_{PWV})$ is regarded as a spatial energy regulator.
            \begin{itemize}
                \item \textbf{Low-Vapour Regions}: As the thermal conditions for precipitation formation are absent, $M_{spatial} \rightarrow 0$. The multiplication operation effectively attenuates the energy of the radar signal, thereby suppressing erroneous echo responses in this region.
                \item \textbf{High-Vapour Regions}: $M_{spatial} \rightarrow 1$. The gate is in the conductive state, permitting the high-frequency textural features of the radar to pass through.
            \end{itemize}
            
            Finally, residual connections prevent gradient vanishing caused by multi-layer gating while preserving the high-frequency details of the original radar signal.
            
            This fusion process is applied independently to both the hidden states (H) and cell states (C), resulting in the final fused states $H_{Fused}$ and $C_{Fused}$.

    \subsection{Recurrent Decoder with Autoregression}
    The decoder is responsible for projecting the fused features back to the pixel-level precipitation intensity. It is initialized with the fused states $H_{Fused}$ and $C_{Fused}$ from the RPF module.
    
    At each prediction step $t$, the decoder takes the prediction from the previous step $\hat{Y}_{t-1}$ as input. The network comprises two upsampling blocks using Deconvolution to restore the spatial resolution.
    This architecture allows the model to generate accurate future radar frames.
    
\section{Experiments}
    \subsection{Data Sources and Baseline}
        \subsubsection{Study Area}
\begin{figure}[htbp]
                \centering
                \includegraphics[width=0.8\linewidth]{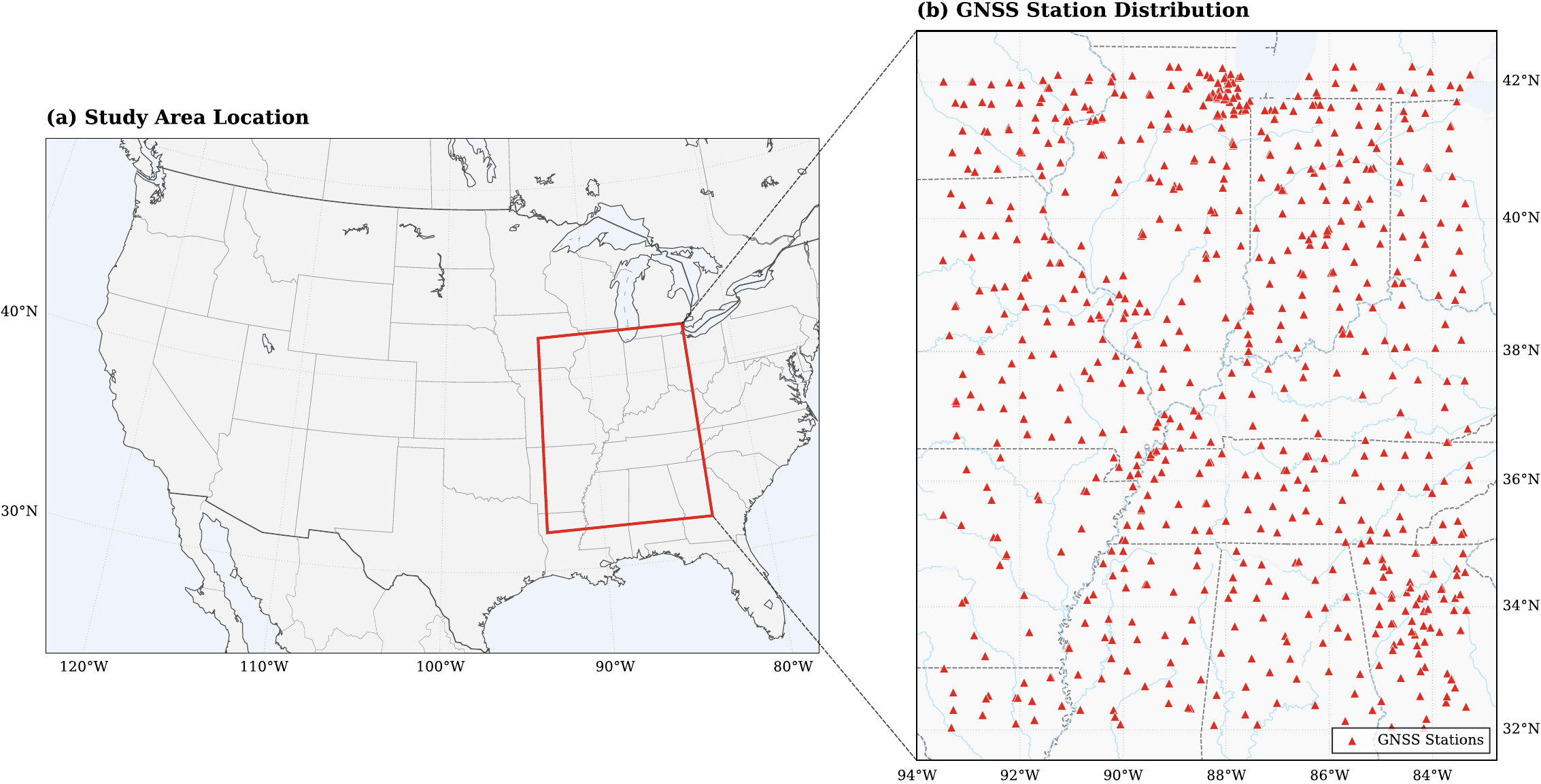}
                \caption{(a) The geographic location of the study area (red bounding box) within the Contiguous United States. (b) Zoomed-in view showing the spatial distribution of the GNSS stations (red triangles) used in this work.}
                \label{fig: study_area}
            \end{figure}

            This study focuses on the MRB, which covers approximately 3.2 million $\text{km}^2$, accounting for 41\% of the land area of the continental United States (Figure~\ref{fig: study_area}). Precipitation in this region is influenced by both mesoscale convective systems and cyclonic activity, resulting in complex hydrological processes. In recent years, both the frequency of floods and the magnitude of flood peaks have increased significantly, making it an ideal study area for evaluating improvements in precipitation nowcasting and its hydrological impacts.
        \subsubsection{PWV}
            PWV is derived from tropospheric products provided by the Nevada Geodetic Laboratory (NGL) at the University of Nevada\cite{gnsspwv}. NGL processes raw GNSS observations from over 17,000 globally distributed stations using the GipsyX software and reprocessed orbit products from the Jet Propulsion Laboratory. 
            Within the MRB domain, \textbf{743 stations} with continuous records longer than 90\% availability are selected (Figure~\ref{fig: study_area}). 
            
\begin{figure}[htbp]
            \centering
            \includegraphics[width=0.9\linewidth]{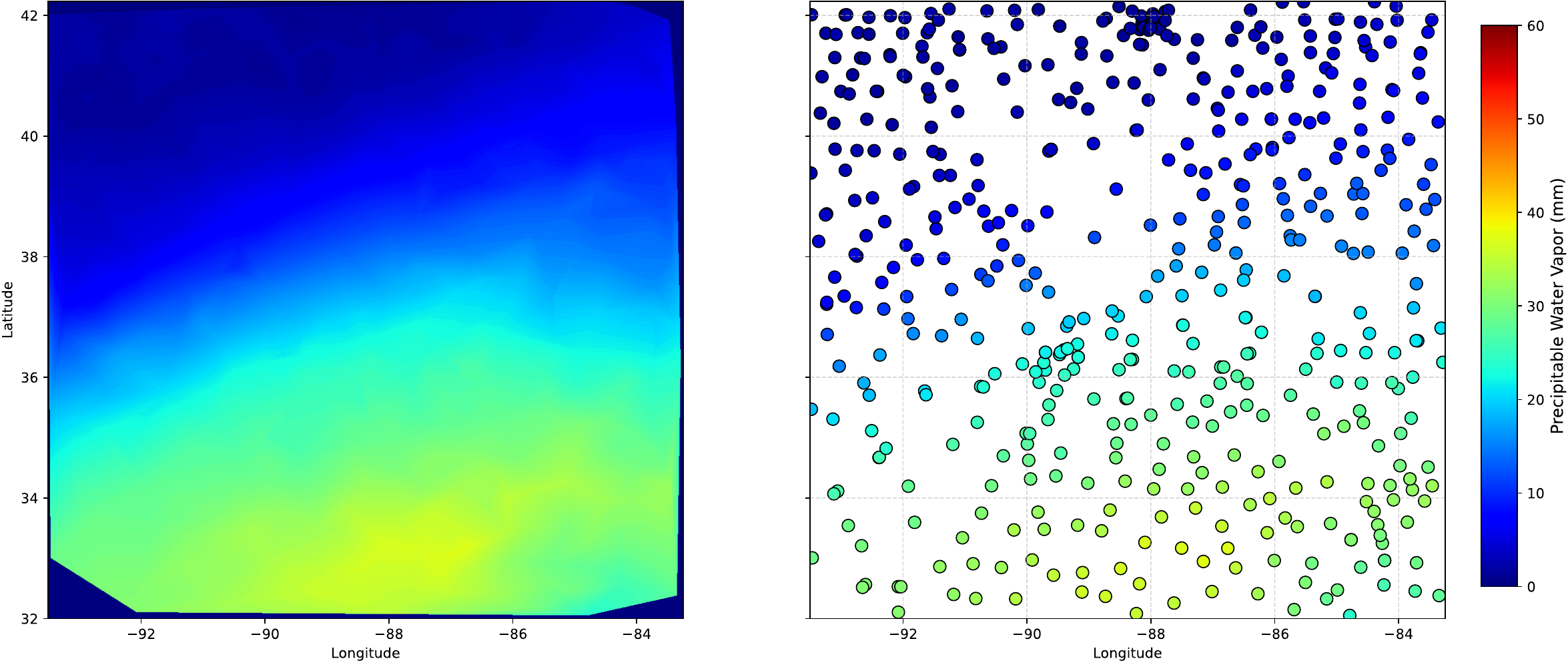}
            \caption{Comparison: Interpolated Image vs. Ground Truth in the PWV Dataset at 2023-02-01 00:00:00 with 496 Valid Stations. (left) Interpolated Image, (right) Ground Truth.}
            \label{fig: pwv_intro}
            \end{figure}

            PWV data frames are extracted and aggregated to a uniform 10-minute temporal resolution to align with the radar data. 
            Due to the sparsity of GNSS stations, PWV values are spatially interpolated to the same 1024 $\times$ 1024 grid using a bilinear method.
            Figure~\ref{fig: pwv_intro} shows the final result of the processing. The final processed PWV dataset spans from \textbf{February 1, 2023, to June 30, 2023}.
        \subsubsection{Radar QPE Products}
            High-resolution radar observations are obtained from the Multi-Radar/Multi-Sensor (MRMS) system developed and operationally maintained by the National Oceanic and Atmospheric Administration (NOAA) National Severe Storms Laboratory (NSSL) and National Centers for Environmental Prediction (NCEP)\cite{MRMS}. The primary product used in this study is the QPE product at a 0.01° ($\sim1km$) horizontal resolution and a 2-minute temporal resolution.
            The processed radar dataset covers a period from \textbf{February 1, 2023, to June 30, 2023}.
            
            To align the dataset with the study domain and model requirements, the following preprocessing workflow was implemented:
            \begin{itemize}
                \item \textbf{Spatial Coverage}: Data are extracted within the MRB bounded by \SIrange{32.0}{42.24}{\degree}\,N, \SIrange{93.49}{83.25}{\degree}\,W.
                \item \textbf{Radar Data Representation}: For each radar QPE frame, it is saved as a 1024 $\times$ 1024 image. 
                \item \textbf{Temporal Sampling}: To balance computational efficiency and align with the original sampling interval of NowcastNet, data are temporally sampled to 10-minute granularity
            \end{itemize}
        
            The final dataset consists of temporally continuous 10-minute frames tailored for sequence prediction tasks.
        \subsubsection{Baselines}
        The methods in our comparative evaluation are listed as follows.
        \begin{itemize}
            \item \textbf{Optical flow}\cite{pysteps}:An open-source benchmark utilizing semi-Lagrangian advection for precipitation extrapolation.precipitation trajectories based on optical flow.
            \item \textbf{ConvLSTM}\cite{convlstm}: A foundational framework integrating convolution into LSTM cells to capture spatiotemporal dependencies. 
            \item \textbf{PredRNN}\cite{wang2022predrnn}: Utilise Spatiotemporal LSTM units to enhance memory flow for capturing complex spatial variations.
            \item \textbf{NowcastNet}\cite{nowcastnet}: A state-of-the-art generative framework that combines physical evolution constraints with conditional GANs to generate sharp and physically consistent precipitation forecasts.
        \end{itemize}
    \subsection{Evaluation Metrics}
    To comprehensively evaluate the performance of the proposed model, we employ two types of metrics: continuous metrics for pixel-level intensity accuracy and categorical metrics for precipitation event detection capability. All metrics are computed over the test dataset.
        \subsubsection{Continuous Metrics}
            We utilise Mean Absolute Error (MAE) and Root Mean Square Error (RMSE) to quantify the global deviation between the predicted precipitation intensity and the ground truth. Let $P$ denote the predicted rainfall intensity and $G$ denote the ground truth value. For a total of $N$ grid points across the spatial domain and temporal sequence, the metrics are defined as follows:
            \begin{equation}
                \text{MAE} = \frac{1}{N} \sum_{i=1}^{N} |P_i - G_i|
            \end{equation}
            \begin{equation}
                \text{RMSE} = \sqrt{\frac{1}{N} \sum_{i=1}^{N} (P_i - G_i)^2}
            \end{equation}
            MAE reflects the average magnitude of errors, while RMSE penalizes larger errors more heavily, making it sensitive to extreme rainfall events.
        \subsubsection{Categorical Metrics}
            To assess the model's performance in forecasting rainfall events at different intensities, we convert the continuous precipitation maps into binary matrices using specific thresholds. Based on the meteorological distribution of the dataset, we selected three thresholds: $\tau \in \{0.1, 1.0, 4.0\}$, representing light, moderate, and heavy rainfall regimes, respectively.
            
            For a given threshold $\tau$, a grid point is classified as a positive event if the value $\ge \tau$. We calculate TP (forecast = 1, truth = 1), FN (forecast = 0, truth = 1), and FP (forecast = 1, truth = 0) , and TN (forecast = 0, truth = 0) to compute the Critical Success Index (CSI): 
            \begin{equation}
                \text{CSI} = \frac{\text{TP}}{\text{TP} + \text{FP} + \text{FN}}
            \end{equation}
            \textbf{CSI} provides a comprehensive score considering both false alarms and missed events (range $[0, 1]$, optimal is 1).
        \subsubsection{Temporal Evaluation}
            Since the reliability of nowcasting typically degrades with increasing lead time, we conduct a frame-by-frame analysis. Specifically, we track these metrics at critical lead times: $T=10min$ , $T=40min$, $T=80min$ , and $T=120min$, to observe the temporal evolution of the model's predictive capability.
    \subsection{Experiment Results}
        \begin{table*}
            \centering
            \begin{tabular}{llrrrr}
                \toprule
                \textbf{Threshold} & \textbf{Model} & \multicolumn{4}{c}{\textbf{CSI}} \\
                \cmidrule(lr){3-6}
                & & \textbf{T=10 min} & \textbf{T=40 min} & \textbf{T=80 min} & \textbf{T=120 min} \\
                \midrule

                $\boldsymbol{\tau=0.1}$
                & OpticalFlow & 0.680 & 0.493 & 0.397 & 0.340 \\
                & ConvLSTM & 0.602 & 0.460 & 0.383 & 0.336 \\
                & PredRNN & 0.365 & 0.136 & 0.130 & 0.109 \\
                & NowcastNet & 0.478 & 0.477 & 0.477 & 0.477 \\
                & \textbf{FusionCast} & \textbf{0.784} & \textbf{0.661} & \textbf{0.614} & \textbf{0.534} \\
                \addlinespace

                $\boldsymbol{\tau=1.0}$
                & OpticalFlow & 0.609 & 0.426 & 0.333 & 0.278 \\
                & ConvLSTM & 0.560 & 0.392 & 0.301 & 0.238 \\
                & PredRNN & 0.572 & 0.120 & 0.109 & 0.092 \\
                & NowcastNet & 0.432 & 0.432 & 0.431 & 0.431 \\
                & \textbf{FusionCast} & \textbf{0.723} & \textbf{0.623} & \textbf{0.588} & \textbf{0.529} \\
                \addlinespace

                $\boldsymbol{\tau=4.0}$
                & OpticalFlow & 0.360 & 0.207 & 0.149 & 0.121 \\
                & ConvLSTM & 0.308 & 0.093 & 0.005 & 0.000 \\
                & PredRNN & 0.276 & 0.029 & 0.023 & 0.023 \\
                & NowcastNet & 0.207 & 0.207 & 0.208 & 0.208 \\
                & \textbf{FusionCast} & \textbf{0.442} & \textbf{0.287} & \textbf{0.265} & \textbf{0.213} \\
                \bottomrule
            \end{tabular}
            \caption{Performance comparison of $\text{CSI} (\uparrow$) scores across different precipitation thresholds ($\tau$) and forecasting horizons. The best results for each column are highlighted in \textbf{bold}.}
            \label{tab:comprehensive_categorical_metrics}
        \end{table*}
    In this section, we will compare the forecasting results of the FusionCast model with other precipitation nowcasting models against the aforementioned metrics, in order to validate the effectiveness of our proposed model.
    Experimental data are divided into a training set (March to May 2023), a validation set (June 2023), and a test set (February 2023).
        \subsubsection{Results and Analysis}
\begin{figure}[htbp]
            \centering
            \includegraphics[width=0.45\textwidth]{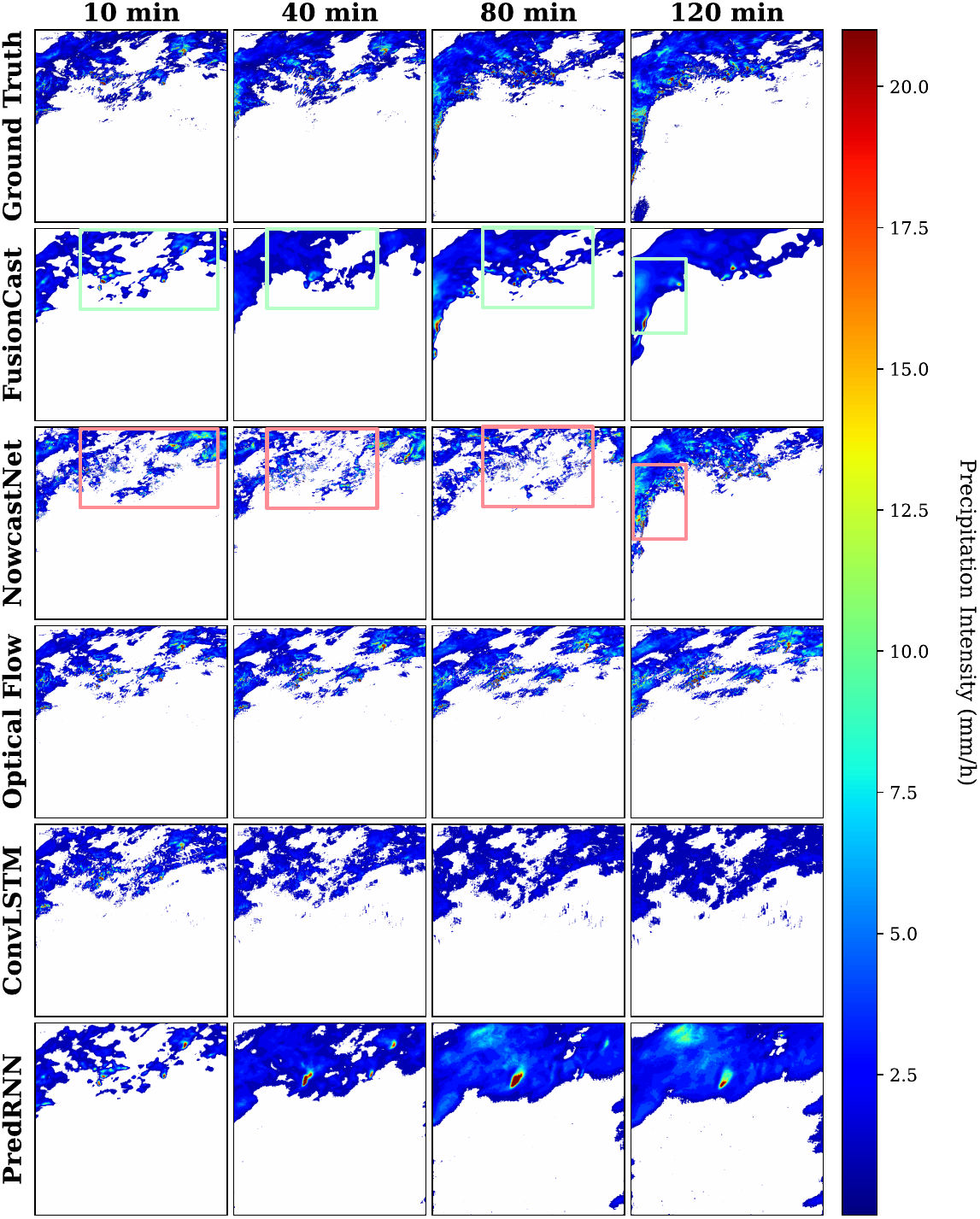}
            \caption{
                Visualisation comparison of precipitation intensity forecast results for the Mississippi River basin over the next two hours, at 12:40 UTC on 22 February 2023. From top to bottom: Ground Truth (GT), FusionCast (Ours), NowcastNet, Optical Flow, ConvLSTM and PredRNN. The highlighted area indicates where our model demonstrates significant improvement over all other models. FusionCast preserves both the spatial structure and intensity of precipitation more consistently, especially at longer lead times.
            } 
            \label{fig: model_compare}
    \end{figure}

        \paragraph{Continuous Metrics Results}
            Table~\ref{tab:continuous_metrics} presents the performance of all models on continuous metrics. 
            FusionCast achieved optimal results for both RMSE and MAE. Significantly, FusionCast demonstrated substantial improvements over the NowcastNet, suggesting that our model effectively minimises overall prediction error. 
            This enhances the accuracy of precipitation intensity forecasts.            
        \paragraph{Classification Performance}
            Table~\ref{tab:comprehensive_categorical_metrics} reports the CSI for multiple models. Results are shown across three precipitation intensity thresholds ($\tau = 0.1, 1, 4~\mathrm{mm/h}$) and lead times from 10 to 120~minutes.
            The figure~\ref{fig: model_compare} illustrates a comparison of the output images from the NowcastNet model and the FusionCast model at different time points.

            As can be seen from the table~\ref{tab:comprehensive_categorical_metrics} and figure~\ref{fig: model_compare}, our model demonstrates a significant improvement over the NowcastNet model in the CSI evaluation metric across all time intervals within the next 120 minutes. The results demonstrate that our model can integrate features from multiple data sources. Leveraging the extracted multimodal data features further improves the prediction results generated by the NowcastNet model
        \begin{table}
                \centering
                \begin{tabular}{lcc}
                    \toprule
                    \textbf{Model}  & \textbf{RMSE (mm/h)} & \textbf{MAE (mm/h)} \\
                    \midrule
                    OpticalFlow& 1.598& 0.214\\
                    ConvLSTM& 1.181& 0.356\\
                    PredRNN& 1.835& 0.743\\
                    NowcastNet& 1.500& 0.179\\
                    \textbf{Ours}& \textbf{1.004}& \textbf{0.103}  \\
                    \bottomrule
                \end{tabular}
                \caption{Comparison of continuous metrics (RMSE, MAE) ($\downarrow$) comparison of precipitation nowcasting performance across the test dataset. The best results are highlighted in \textbf{bold}.}
                \label{tab:continuous_metrics}
        \end{table}
    \subsection{Ablation Study}
    To validate the core components of the FusionCast framework, we conducted ablation studies on multimodal data sources and the RPF fusion module. We focus on the CSI at precipitation thresholds of 0.1, 1, and 4~\si{\mm\per\hour}, as CSI provides a comprehensive measure for convective precipitation nowcasting.
    \subsubsection{Data Source Ablation}
    This experiment evaluates the improvement provided by the sparse \textbf{PWV} field ($X_{GNSS}$) and the \textbf{NowcastNet prior knowledge} ($X_{Radar}^{prior}$).
    \begin{itemize}
        \item \textbf{w/o PWV}: The PWV branch is removed; the model only uses historical radar and future prior knowledge.
        \item \textbf{w/o Prior}: The future prior knowledge branch is removed; the model only uses historical radar and PWV.
    \end{itemize}
    \begin{figure}[htbp]
    \centering
    \includegraphics[width=0.8\linewidth]{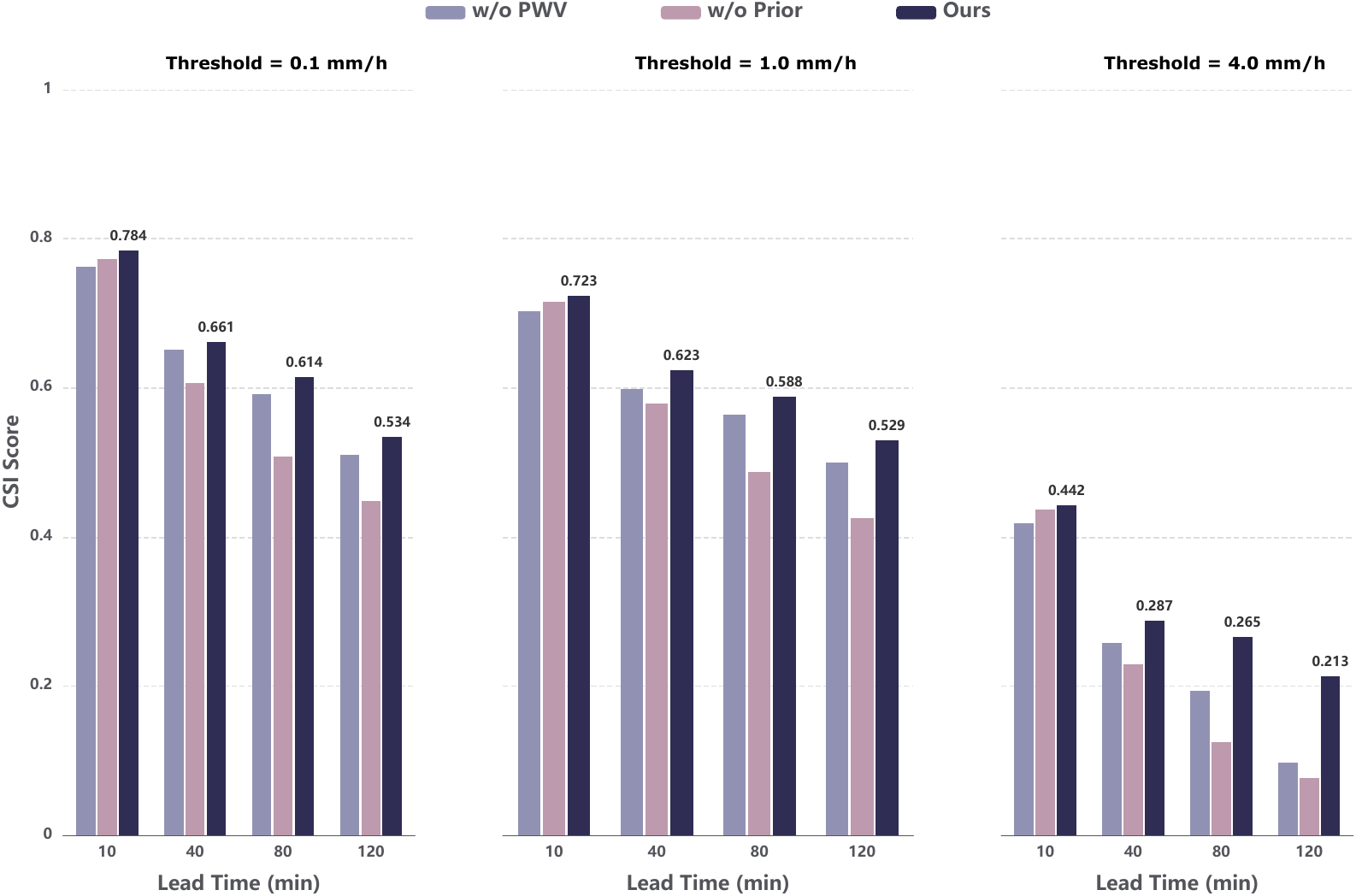}
    \caption{
        Ablation study across different thresholds ($\tau$) and lead times. The inferior performance of "w/o PWV" and "w/o Prior" variants validates that integrating multimodal priors is indispensable for accurate precipitation nowcasting.
    }\label{fig: variant_abliation}
    \end{figure}
%    \paragraph{Analysis}
    Through Figure~\ref{fig: variant_abliation}, the role played by each data source in the model prediction is analysed as follows:
    \begin{itemize}
        \item \textbf{Role of PWV}: Excluding PWV data from FusionCast's input leads to a substantial decrease in the CSI metric of its forecast outputs at different precipitation intensity thresholds compared to the full model. Furthermore, forecast accuracy diminishes as the prediction lead time increases. This suggests that PWV data plays a vital part in the formation, development and evolution of extreme weather events. It can positively influence future rainfall forecasting and enhance the model's long-term stability.
        \item \textbf{Role of Future Prior Knowledge}: The CSI metric for forecasts derived from future prior knowledge outputs based on NowcastNet declines more sharply than when PWV is removed. This suggests that features derived from future prior knowledge are more effective than PWV in predicting future rainfall intensity. It also demonstrates that the prior frame provides explicit \textbf{trajectory guidance} for the future evolution and potential development of precipitation systems, substantially enhancing the model's trajectory accuracy and predictive credibility.
    \end{itemize}
    
    \subsubsection{RPF Module Ablation}
    This experiment validates the effectiveness of the proposed \textbf{RPF Module}.
    \begin{itemize}
        \item \textbf{w/o RPF}: The RPF module is replaced by simple \textbf{channel concatenation} to merge the $H$ and $C$ states from the encoder outputs before initialization of the decoder.
        \item \textbf{Our (Concat)}: Features from the attention outputs are fused via simple \textbf{concatenation}, replacing the gating mechanism.
    \end{itemize}
    \begin{figure}[htbp]
    \centering
    \includegraphics[width=0.8\linewidth]{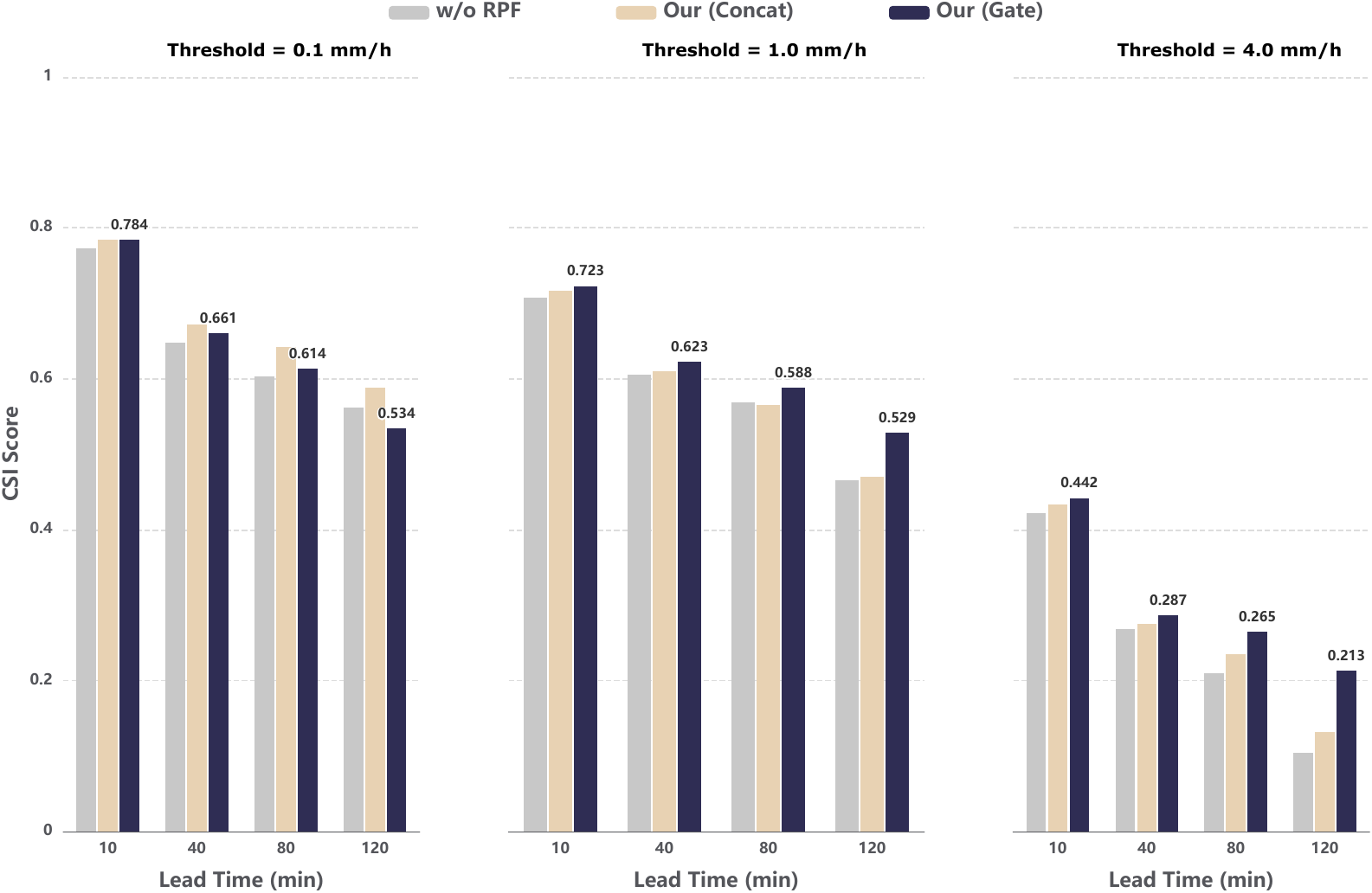}
    \caption{
        The proposed Gating strategy effectively mitigates the feature blurring observed in the Concatenation baseline, leading to superior robustness in forecasting heavy precipitation ($\tau=4.0$ mm/h) over long lead times.
    } 
    \label{fig: rpf_ablation}
    \end{figure}
    % \paragraph{Analysis}
    Based on the comparative results presented in Figure~\ref{fig: rpf_ablation}, the analysis of the role played by the RPF module in model prediction is as follows:
        \begin{itemize}
            \item \textbf{Effectiveness of the RPF Module}: By comparing the FusionCast (Gate) with the FusionCast (w/o RPF), consistent performance improvements are observed across all precipitation thresholds and forecast lead times. Notably, the CSI score indicates a substantial improvement in the 120-minute forecast for heavy precipitation ($\tau = 4.0$ mm/h), confirming the effectiveness of the RPF module in enhancing prediction accuracy.
            \item \textbf{Superiority of the Gating Mechanism}: We compare FusionCast (Gate) with FusionCast (Concat). Significant deterioration in FusionCast (Concat) performance is observed in regions with heavy precipitation ($\tau=4.0$ mm/h). The result demonstrates that our gating mechanism effectively utilises thermal energy to enhance the textural features, thereby ensuring the physical consistency across different modalities. Gating mechanism ensures superior performance for heavy rainfall compared to the concatenation baseline.
        \end{itemize}

\section{Conclusion}
    In this paper, we propose FusionCast to address the limitations of fusion strategies like simple concatenation in precipitation nowcasting. Unlike existing methods, FusionCast effectively integrates multimodal data within a unified framework. Specifically, we designed a Future Prior Processing Module to leverage predictive guidance and a RPF module to handle modality heterogeneity. By utilising a gating mechanism, the RPF module can adaptively align features from different modalities.
    
    Extensive experiments demonstrate that FusionCast significantly outperforms current state-of-the-art benchmark models, validating the effectiveness of differentiated modal processing and the application of future a priori information in precipitation nowcasting.
\bibliographystyle{named}
\bibliography{ijcai26}

\end{document}